\documentclass[lettersize,journal]{IEEEtran}
\usepackage{amsmath,amsfonts}
\usepackage{algorithmic}
\usepackage{array}
\usepackage[caption=false,font=footnotesize]{subfig}
\usepackage{textcomp}
\usepackage{stfloats}
\usepackage{url}
\usepackage{verbatim}
\usepackage{graphicx}
\def\BibTeX{{\rm B\kern-.05em{\sc i\kern-.025em b}\kern-.08em
    T\kern-.1667em\lower.7ex\hbox{E}\kern-.125emX}}
\usepackage{balance}
\usepackage{hyperref}
\usepackage{booktabs}
\usepackage{multirow}
\usepackage{tabularx}
\newcolumntype{C}[1]{>{\centering\arraybackslash}p{#1}}

\usepackage[table]{xcolor}
\hypersetup{hidelinks}

\begin{document}

\title{Geometry-Aligned Semantic Matching for Cross-Modal \\
        Planar Image Registration}

\author{
Zhiwei Wang,
Defeng He$^{*}$,
Yuxing Li,
Meilu Zhu,
Edmund Y. Lam$^{*}$,~\IEEEmembership{Fellow,~IEEE}

\thanks{$*$ Corresponding authors.}%
}

\maketitle

\begin{abstract}
Cross-modal image matching establishes stable and accurate geometric correspondences across modalities for planar registration. Existing semantic representations provide cross-modal consistency, but semantic similarity does not necessarily imply geometric correspondence. Meanwhile, fine-grained CNN features provide accurate local details but lack global cross-modal semantic guidance for stable refinement.
To address these issues, we propose CDPM, which first establishes geometrically consistent semantic representations and then preserves their dominant role in correspondence estimation during fine-grained localization. Specifically, we progressively adapt DINOv3 using geometrically consistent cross-modal patch pairs, enabling feature similarity to better reflect true cross-modal spatial correspondences. We then construct a DINO-Centric Feature Pyramid, where multi-scale DINO representations maintain stable cross-modal correspondences, while a lightweight CNN branch provides auxiliary structural details for precise local refinement.
Extensive experiments on three cross-modal datasets demonstrate the superior performance of CDPM. On VIS-IR, compared with the dense matcher RoMa, CDPM improves AUC@3/5/10/20 by 7.36, 13.40, 13.75, and 10.42 percentage points, respectively, and reduces mACE from 5.83 to 2.78 pixels. It also outperforms RoMa v2 across all metrics while requiring 45.6\% fewer FLOPs.
The online demo and dataset are available, and the code will be released on our project page at \url{https://warren-wzw.github.io/CDPM/}.
\end{abstract}


\section{Introduction}\label{sec1}
\begin{figure}[t]
    \centering

    \subfloat[Motivations of the proposed method.\label{Motivation}]{
        \includegraphics[width=\columnwidth]{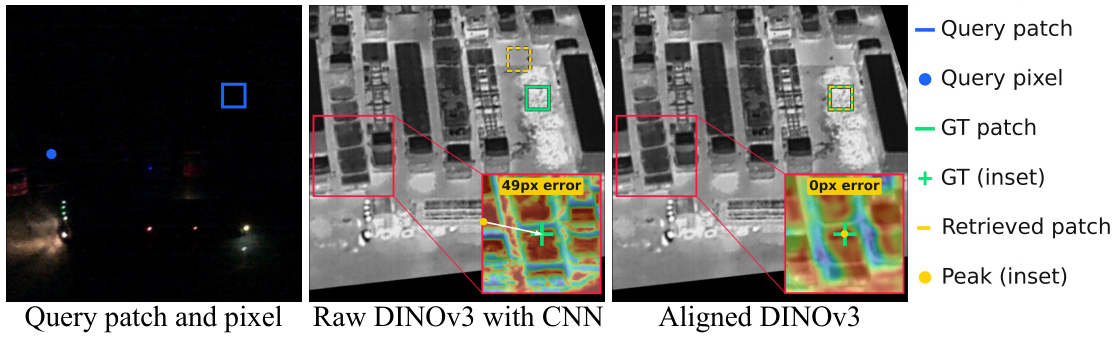}
    }
    \par\vspace{-0.35cm}
    \subfloat[Qualitative comparison of correspondence and alignment.\label{Teaser}]{
        \includegraphics[width=\columnwidth]{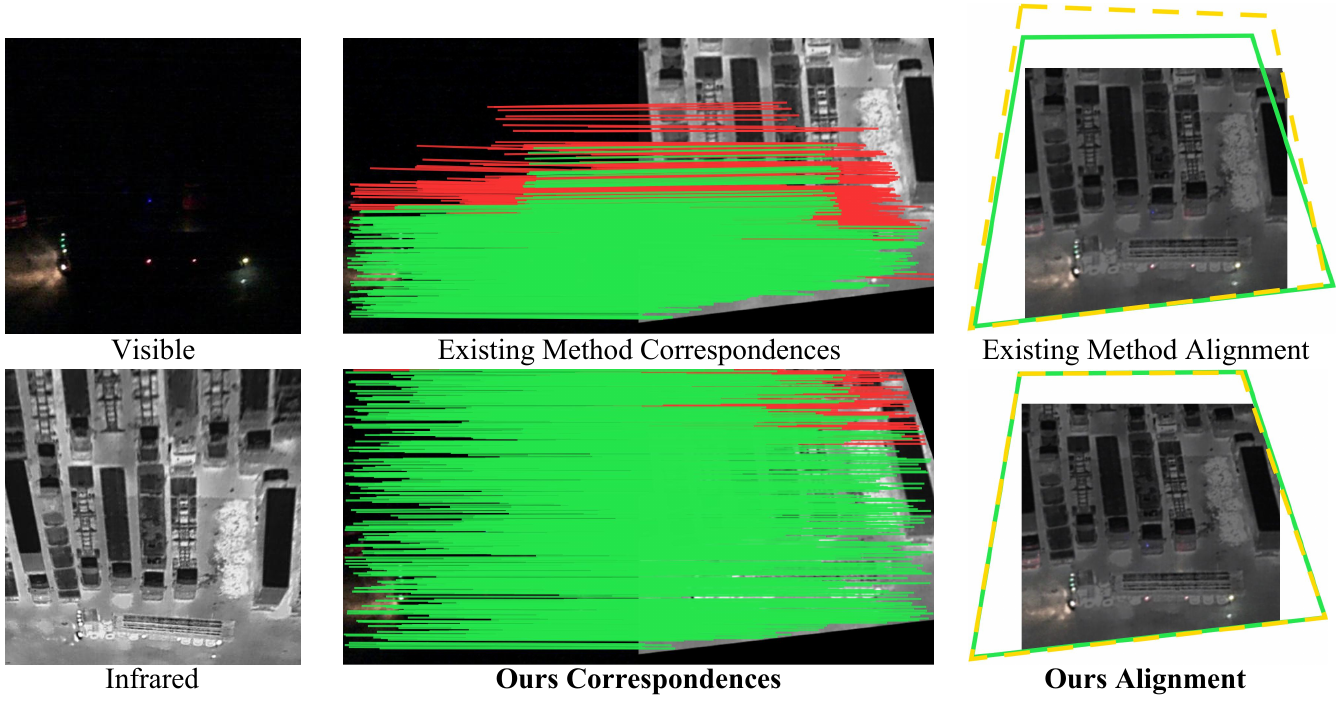}
    }

    \caption{(a) Raw DINOv3 yields ambiguous cross-modal responses, while CNN refinement may sharpen but shift the peak; CDPM aligns semantic correspondence and stabilizes fine-scale refinement.
    (b) Green and red correspondences indicate reprojection errors of $\leq 3$ and $>3$ pixels, respectively. Yellow dashed boundaries denote the predicted alignments.}
    \label{fig:teaser_motivation}
    \vspace{-0.6cm}
\end{figure}
\IEEEPARstart{C}{ross-modal} image registration aims to geometrically align images captured by different sensing modalities~\cite{zheng2025plug}. Reliable dense correspondences provide an important foundation for image fusion~\cite{TANG2023PSFusion,yi2024diff,liu2026cl}, multimodal perception~\cite{liu2022target,Wang2025DiFusionSegDS}, and dense 3D reconstruction~\cite{yao2018mvsnet, guo2025multi}. For approximately planar scenes, reliable correspondences can be further used with robust estimators to recover the homography relating the two images~\cite{sun2021loftr,edstedt2024roma}. However, substantial differences in imaging mechanisms, texture and intensity across modalities make reliable pixel-level correspondences difficult to establish~\cite{jiang2021review,ren2025minima}. Therefore, the difficulty lies in balancing correspondence stability and localization accuracy. Robust cross-modal representations are often too coarse for localization, whereas fine-grained features tend to be modality-sensitive.

Existing approaches to planar image registration generally include direct homography estimation and correspondence-based matching. Detector-free and dense matchers progressively refine correspondence estimates through multi-scale feature representations and coarse-to-fine refinement~\cite{edstedt2023dkm,wang2024eloftr}. More recently, vision foundation models have been introduced to provide robust semantic priors for image matching and geometric estimation~\cite{zhang2025adapting,jiang2024omniglue,wang2025tex2sem}. Representative dense matchers combine pretrained DINO features for coarse correspondence estimation with CNN features for fine-scale localization~\cite{edstedt2024roma,edstedt2025roma}. This combination of foundation-model semantics and CNN details provides an effective paradigm for same-modality matching, where semantic representations and local structures remain relatively consistent across views. Extending this paradigm to cross-modal matching, however, requires semantic representations to be aligned across modalities while preserving reliable fine-scale cues, giving rise to two key challenges in practical settings.

First, cross-modal semantic alignment does not necessarily ensure geometric correspondence. Although large-scale pretraining provides strong semantic representations and cross-domain generalization~\cite{he2022masked,simeoni2025dinov3}, its objectives primarily emphasize semantic consistency and appearance invariance rather than cross-modal geometric correspondence. Recent studies further adapt foundation-model representations across heterogeneous modalities to improve cross-modal consistency and transferability~\cite{Wei_2026_CVPR,Kabra_2026_CVPR}. However, such alignment is not explicitly optimized for registration-specific geometric correspondence. Consequently, the most similar features may not correspond to the same physical location across modalities. As shown in~\autoref{Motivation}, raw DINOv3 produces high responses near the ground-truth correspondence, yet the correct patch does not necessarily achieve the highest similarity among all candidates. Such ambiguities become more pronounced under repetitive structures, low-light conditions, and substantial modality gaps in practical registration scenarios.

Second, fine-scale refinement may compromise cross-modal correspondence stability. High-resolution CNN features provide fine spatial details for precise localization and are widely used in coarse-to-fine matching and local refinement~\cite{sun2021loftr,edstedt2024roma,wang2024eloftr}. However, they are also highly responsive to local textures, edges, and intensity variations that can differ substantially across modalities. Without sufficient and consistent cross-modal semantic guidance, these modality-sensitive features may favor locally salient but geometrically incorrect responses. As shown in~\autoref{Motivation}, CNN features produce sharper matching responses, but their peaks may deviate from the ground-truth correspondence. Consequently, increasing reliance on such features throughout iterative refinement can gradually pull otherwise reliable coarse correspondences toward incorrect local responses, leading to refinement drift and degraded localization accuracy.

To address these issues, we propose CDPM (Cross-modal Dense Pyramid Matcher), a dense matching framework for cross-modal planar registration. CDPM first adapts DINOv3 using geometric correspondence supervision and then leverages the adapted semantic features to guide fine-scale localization.
As shown in~\autoref{Teaser}, CDPM establishes more reliable and geometrically consistent cross-modal correspondences, enabling accurate planar alignment.
Specifically, we construct geometrically consistent positive and negative cross-modal patch pairs and perform staged adaptation of DINOv3, enabling feature similarity to more accurately reflect true spatial correspondences. On this basis, we develop a DINO-Centric Feature Pyramid (DCFP), where adapted multi-level DINO representations are preserved throughout the feature pyramid. They provide the primary correspondence cues across scales, while a lightweight CNN branch supplements local spatial details. By maintaining DINO semantics as the dominant cue for cross-modal correspondence, CDPM preserves stable matching during refinement. CNN features are introduced only as auxiliary cues to improve localization accuracy. 
The main contributions are summarized as follows:
\begin{itemize}
    \item We propose a patch-level cross-modal adaptation strategy that uses homography-guided patch pairs and staged contrastive learning to align pretrained DINO representations with geometric correspondence.

    \item We propose DCFP, where adapted multi-level DINO features dominate correspondence estimation, while a lightweight CNN selectively supplements local structural details as auxiliary cues at finer scales to improve matching stability and localization accuracy.

    \item We construct PCB-Layout, an industrial registration dataset comprising PCB images and design layouts. Experiments on two public datasets and PCB-Layout demonstrate the superior registration performance of CDPM across diverse cross-modal settings.
\end{itemize}
\begin{figure*}[!t]
    \centering
    \includegraphics[width=1.0\textwidth]{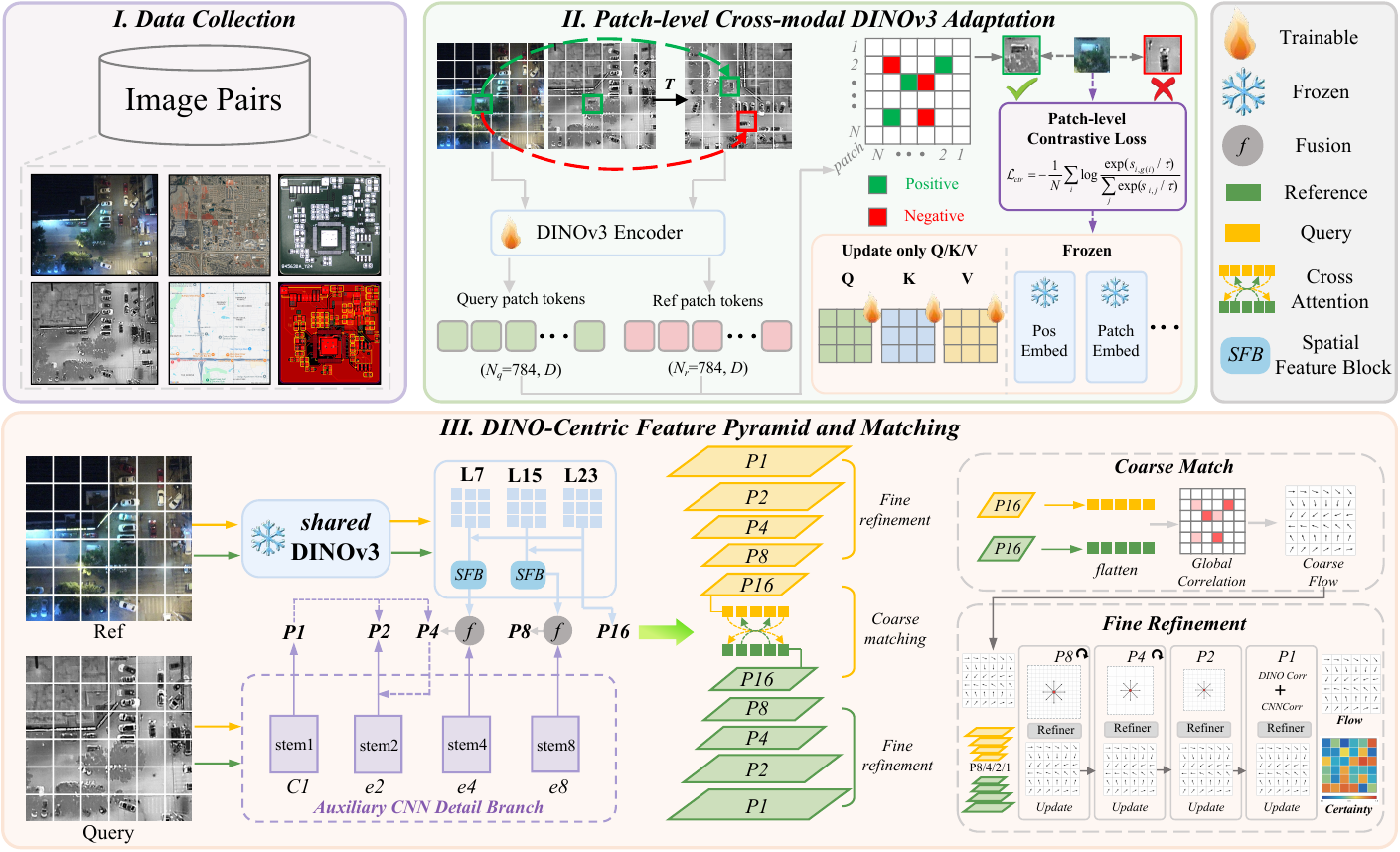}
    \caption{Overview of the proposed CDPM, including data collection, patch-level cross-modal DINOv3 adaptation, and DINO-centric feature pyramid construction that preserves DINO correspondence representations throughout coarse-to-fine matching.}
    \label{ModelArch}
    \vspace{-0.6cm}
\end{figure*}
\section{Related Work}\label{sec2}
\subsection{Cross-modal Homography Estimation}
Cross-modal planar image registration aims to estimate the spatial transformation between images captured in different modalities~\cite{wei2026grid}. Traditional methods typically extract carefully designed structural features robust to radiometric variations and recover geometric models from sparse correspondences. RIFT~\cite{li2019rift} detects features using phase congruency and constructs descriptors from multi-orientation Log-Gabor responses to alleviate nonlinear radiometric differences. With the development of deep learning, learning-based methods directly estimate homographies through cross-image feature interactions. RHWF~\cite{cao2023recurrent} progressively refines the transformation via homography-guided image warping and recurrent prediction, while MCNet~\cite{zhu2024mcnet} further employs multi-scale correlation search to balance accuracy and efficiency. Unlike these direct homography estimation methods, GFNet~\cite{zhang2025adapting} performs dense matching on regular grids to preserve high-resolution information while maintaining computational efficiency. However, under substantial modality gaps in practice, homography estimation remains limited by the reliability and localization accuracy of cross-modal correspondences. Therefore, CDPM focuses on establishing robust and accurate dense cross-modal correspondences, from which the homography is recovered through robust estimation.

\subsection{Detector-Free and Dense Matching}
Detector-free matching methods establish correspondences directly on regular feature grids, avoiding the instability of keypoint detection in low-texture and repetitive regions. LoFTR~\cite{sun2021loftr} employs self- and cross-attention to establish globally consistent semi-dense correspondences, followed by local matching for fine localization. ELoFTR~\cite{wang2024eloftr} reduces feature interaction costs through aggregated attention and adaptive token selection, while introducing two-stage correlation refinement for improved localization accuracy. SceneGlue~\cite{du2026sceneglue} further incorporates scene-level context and cross-view visibility into feature matching. XoFTR~\cite{tuzcuouglu2024xoftr} further enhances cross-modal feature consistency through masked image modeling and pseudo-thermal augmentation. Unlike semi-dense matching, DKM~\cite{edstedt2023dkm} formulates matching as dense correspondence field estimation and progressively recovers pixel-level correspondences through multi-scale residual updates. These methods have advanced image matching toward coarse-to-fine dense prediction. However, under substantial modality gaps, modality-sensitive fine-scale features may disturb reliable correspondences established at coarser scales, leading to refinement drift. CDPM therefore focuses on preserving cross-modal correspondence stability  across different scales throughout coarse-to-fine refinement.

\subsection{Generalizable Cross-modal Image Matching}
The DINO series of visual foundation models has acquired strong semantic generalization capabilities through large-scale self-supervised learning~\cite{simeoni2025dinov3,oquab2023dinov2}. OmniGlue~\cite{jiang2024omniglue} and RoMa~\cite{edstedt2024roma} employ frozen DINOv2 features for robust coarse matching and combine them with specialized CNN features for fine localization, while RoMa v2~\cite{edstedt2025roma} further incorporates DINOv3 and multi-view feature interaction to improve general-purpose dense matching. Another line of work focuses on scaling training data. MINIMA~\cite{ren2025minima} constructs large-scale synthetic cross-modal data using generative models, AnyMatch~\cite{yang2026anymatch} generates geometrically consistent multi-modal pairs from single-view images via 3D reprojection, while MatchAnything~\cite{he2025matchanything} learns transferable matching capabilities from diverse data sources and cross-modal supervision. Despite the strong generalization of foundation-model features, their generic semantic representations are not explicitly aligned with cross-modal geometric correspondence, such that semantic similarity may not reliably indicate the correct physical correspondence.

\section{Methodology}\label{sec3}
\subsection{Problem Formulation and Overview}
Given a cross-modal image pair $(I_q, I_r)$ depicting the same planar scene, our goal is to estimate a homography matrix $\mathbf{H}$ that warps the query image $I_q$ into alignment with the reference image $I_r$. For a point $\mathbf{x}_q$ in the query image, its corresponding point $\mathbf{x}_r$ in the reference image is given by:
\begin{equation}
\mathbf{x}_r
=
\pi\!\left(\mathbf{H}\tilde{\mathbf{x}}_q\right),
\qquad
\pi\!\left([u,v,w]^\top\right)
=
[u/w,v/w]^\top,
\end{equation}
where $\tilde{\mathbf{x}}_q=[x_q,y_q,1]^\top$ denotes the homogeneous coordinate of $\mathbf{x}_q$, and $\pi(\cdot)$ denotes the dehomogenization operator.

Rather than directly regressing the global homography, we first extract multi-scale features from both images to predict a dense correspondence field and its associated confidence map. Reliable correspondences are then selected to recover the homography matrix $\mathbf{H}$ through robust estimation. The main challenge of cross-modal matching lies in balancing correspondence stability and localization accuracy. Three factors limit this balance in practice. First, cross-modal appearance variations cause geometrically corresponding regions to diverge in feature space, making feature similarity an unreliable indicator of true spatial correspondence. Second, deep semantic features provide robust correspondence cues but have limited spatial precision, restricting localization accuracy at the patch level. Third, modality-sensitive fine-scale features may disturb reliable coarse correspondences during iterative refinement, causing them to drift toward incorrect local responses, which we refer to as \emph{refinement drift}.

The overall framework of the proposed CDPM is illustrated in~\autoref{ModelArch}. CDPM consists of three stages: patch-level cross-modal DINOv3 adaptation, DINO-Centric feature pyramid construction, and coarse-to-fine matching. First, geometrically consistent cross-modal patch correspondences are constructed and employed as supervision to adapt DINOv3~\cite{simeoni2025dinov3} to cross-modal matching. By pulling the feature representations of geometrically corresponding tokens from different modalities closer together, this process improves cross-modal consistency and discriminability while preserving the semantic generalization capability inherited from pretraining, thereby reducing representation discrepancy and making corresponding regions identifiable at the coarse scale. Next, we construct DCFP, where multi-level DINO representations are progressively reorganized and propagated toward higher resolutions to preserve stable cross-modal correspondences. An auxiliary CNN branch supplements high-frequency structural cues for precise localization. This DINO-centric construction propagates coarse-scale correspondence reliability to higher resolutions, alleviating localization error while maintaining semantic consistency. Finally, global correspondences are established at the coarse scale and progressively refined through residual updates. By preserving DINO correspondence cues and limiting iterative updates at modality-sensitive fine scales, CDPM suppresses refinement drift while progressively improving localization accuracy. The model ultimately outputs a dense correspondence field and its associated confidence map, from which reliable matches are selected to robustly recover $\mathbf{H}$.

\subsection{Patch-level DINOv3 Adaptation}
DINOv3~\cite{simeoni2025dinov3}, pretrained on large-scale visual datasets, exhibits strong semantic generalization capabilities. However, under inputs from different modalities, patches corresponding to the same scene location may still produce substantially different feature representations, making it difficult to obtain distinctive correlation responses at the true correspondence locations. As shown in~\autoref{Motivation}, the original DINOv3 struggles to accurately identify cross-modal corresponding patches. To address this issue, we adapt its patch representations using the geometric correspondences between cross-modal images, thereby enhancing the cross-modal invariance and geometric discriminability of the learned features.

For spatially aligned image pairs, directly treating tokens at the same grid locations as positive pairs can create a positional shortcut. Since the positional embeddings remain fixed, the model may exploit the shared token indices as a direct correspondence cue, rather than learning genuine cross-modal content consistency. To avoid this positional shortcut, we randomly sample a constrained homography \(\mathrm{\mathbf{H}}_{aug}\) and apply it to one of the input images. The randomly sampled homography places the same content patch at different absolute token positions across training samples and induces spatially varying displacements. This breaks the deterministic association between fixed token indices and geometric correspondences, thereby forcing the model to establish correspondences based on cross-modal content rather than positional embeddings.

For the $i$-th patch in the query image, its corresponding position in the reference image is given by $\mathbf{p}_i^r=\pi\!\left(\mathbf{H}_{aug}\tilde{\mathbf{p}}_i^q\right)$, where $\tilde{\mathbf{p}}_i^q$ denotes the homogeneous coordinate of the patch center. The corresponding token is selected according to the mapped position, while patches mapped outside the valid image region are discarded. Geometrically corresponding tokens are treated as positive pairs, whereas tokens at other locations are regarded as negative samples. A patch-level contrastive loss is then employed to make geometrically corresponding tokens reliably identifiable across modalities.

The proposed DCFP integrates representations from different ViT~\cite{dosovitskiy2021image} layers. Since shallow features primarily encode modality-sensitive textures and edges~\cite{ranftl2021vision}, we preserve these pretrained representations and progressively adapt the intermediate and high-level layers for cross-modal feature alignment in two stages: the first stage optimizes Layers 8--15 to establish cross-modal structural correspondences, while the second stage further optimizes Layers 16--23 to propagate such consistency into high-level semantic representations. To preserve the pretrained priors of DINO and reduce the training cost, only the $Q/K/V$ projection matrices in the multi-head self-attention modules are updated at each stage, while all remaining parameters are kept frozen. Both stages are optimized using the aforementioned patch-level contrastive loss, and the adaptation process is formulated as follows:
\begin{equation}
    \small
    \mathcal{T}^{(s)}=
    \begin{cases}
    \left\{\mathbf{W}_{l}^{QKV}\right\}_{l=8}^{15},  & s=1,\\[1.5pt]
    \left\{\mathbf{W}_{l}^{QKV}\right\}_{l=16}^{23}, & s=2,
    \end{cases}
\end{equation}
where $\mathcal{T}^{(s)}$ denotes the set of trainable parameters at stage $s$, and $\mathbf{W}_{l}^{QKV}$ represents the $Q/K/V$ projection matrices in the $l$-th ViT layer. After Stage 1, Layers 8--15 are frozen, and only Layers 16--23 are updated in Stage 2. The adapted DINOv3 backbone is then kept frozen during downstream matching-network training.

\subsection{DINO-Centric Feature Pyramid}
Adapted DINO features provide robust cross-modal correspondences but lack spatial precision, whereas CNN features capture fine structures but remain sensitive to modality-specific appearances. We therefore construct a DINO-Centric Feature Pyramid (DCFP), where multi-level DINO representations serve as the primary correspondence features across scales, while lightweight CNN features are selectively incorporated as auxiliary structural cues. As illustrated in Part~III of \autoref{ModelArch}, DINO correspondences are progressively propagated from coarse to fine scales, with CNN details supplementing local spatial information for precise localization.

\subsubsection{Multi-level DINO Pyramid Construction}
Although the adapted DINO features establish reliable cross-modal correspondences, their final-layer tokens remain constrained by a patch resolution, limiting fine-grained localization. Naive upsampling only enlarges feature maps without introducing new spatial details. Inspired by DPT~\cite{ranftl2021vision}, we extract features from multiple DINO layers and use Spatial Feature Block (SFB) to reassemble them into a coarse-to-fine feature hierarchy.

We extract multi-level features from the 7th, 15th, and 23rd layers of the shared DINO backbone, denoted by $\{\mathbf{Z}_l^v\}_{l\in\{7,15,23\}}$, where $v\in\{q,r\}$. Since the L23 features provide the highest-level semantic representation, they are first used to establish cross-view correspondences at the coarsest P16 scale. Specifically, the features are projected into a unified embedding space and augmented with positional encoding:
\begin{equation}
\widehat{\mathbf Z}_{23}^{v}
=
E_{pos}\left(
T_c(\mathbf Z_{23}^{v})
\right),
\end{equation}
where $T_c(\cdot)$ denotes a $1\times1$ channel projection and $E_{pos}(\cdot)$ denotes the positional encoding.

The projected features are then fused through a bidirectional cross-attention module to obtain the cross-view enhanced P16 features for robust matching:
\begin{equation}
\mathbf P_{16}^{v}=\widehat{\mathbf Z}_{23}^{v}+\operatorname{CA}\left(\widehat{\mathbf Z}_{23}^{v},\widehat{\mathbf Z}_{23}^{\bar v}\right),
\end{equation}
where $v\in\{q,r\}$, $\bar v$ denotes the opposite view, $\operatorname{CA}(\cdot,\cdot)$ denotes cross-attention.

\begin{figure}[t]
    \centering
    \includegraphics[width=0.48\textwidth]{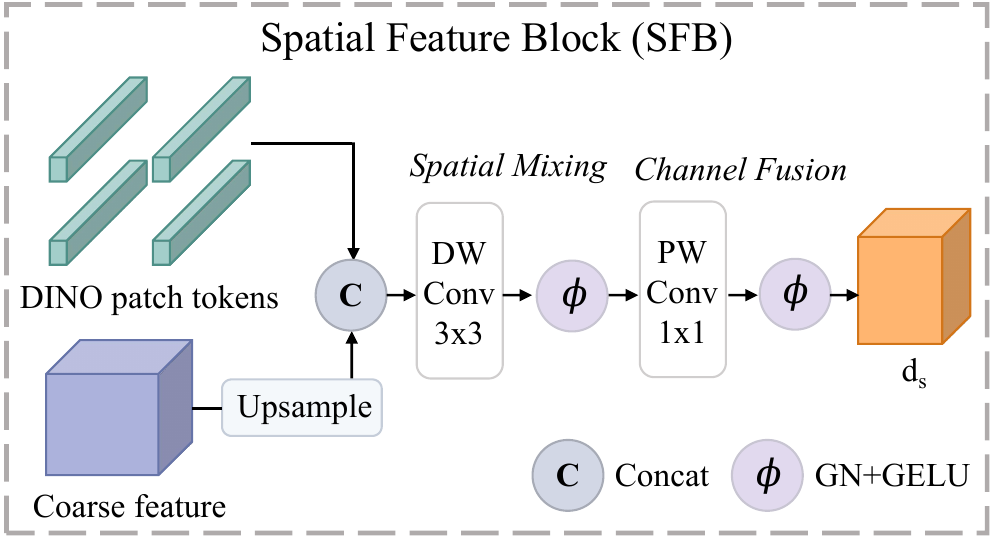}
    \caption{Structure of the proposed Spatial Feature Block (SFB) for progressive multi-level DINO feature aggregation.}
    \label{SFB}
    \vspace{-0.6cm}
\end{figure}

To progressively recover spatial resolution from DINO features, we first project intermediate features into a shared channel space. We denote the channel-projected multi-level DINO features as $\mathbf{Q}_l^v = T_l(\mathbf{Z}_l^v)$, where $l\in\{7,15,23\}$. As illustrated in~\autoref{SFB}, at the P8 and P4 scales, the SFB upsamples the preceding coarse feature, concatenates it with a scale-aligned DINO feature, and processes them using a depthwise-separable convolution:
\begin{equation}
\begin{aligned}
&\operatorname{SFB}_s(\mathbf a_s,\mathbf b_s)=\\[-1pt]
&\qquad
\phi\Big(
\operatorname{Conv}_{\mathrm{PW}}\big(
\phi\big(
\operatorname{Conv}_{\mathrm{DW}}
([\mathcal{U}_2(\mathbf a_s);\mathbf b_s])
\big)\big)
\Big),
\end{aligned}
\end{equation}
where $\mathbf a_s$ denotes the feature propagated from the preceding coarser pyramid level after scale alignment, $[\cdot;\cdot]$ denotes channel-wise concatenation, $\mathcal U_k(\cdot)$ denotes bilinear upsampling by a factor of $k$, and $\phi$ represents GroupNorm followed by a GELU activation function.

At the P8 scale, the projected L23 feature $\mathbf Q_{23}^{v}$ is first processed by a depthwise-separable convolution to initialize the feature hierarchy. The resulting coarse feature is then fused with the scale-aligned L15 feature through $\operatorname{SFB}_{8}$:
\begin{equation}
    \begin{aligned}
    &\mathbf D_{16}^{v}=\operatorname{DSConv}_{16}\left(\mathbf Q_{23}^{v}\right),\\
    &\mathbf D_{8}^{v}=\operatorname{SFB}_{8}\left(\mathbf D_{16}^{v},\mathcal U_2(\mathbf Q_{15}^{v}))\right..
    \end{aligned}
\end{equation}

The DINO feature $\mathbf D_{8}^{v}$ combines the high-level semantics of L23 with the intermediate spatial structures of L15. At the P4 scale, we further incorporate the spatially richer L7 feature. Since the projected L7 feature $\mathbf Q_{7}^{v}$ remains on the P16 token grid, it is upsampled by a factor of four and fused with the P8-scale DINO feature through $\operatorname{SFB}_{4}$. 
\begin{equation}
    \begin{aligned}
    &\mathbf D_{4}^{v}=\operatorname{SFB}_{4}
    \left(\mathbf D_{8}^{v},\mathcal U_{4}(\mathbf Q_{7}^{v})\right).
    \end{aligned}
\end{equation}

The L7 feature further enriches the spatial structure of the DINO representation. At the P2 scale, the aggregated P4 DINO representation is propagated to higher resolution and processed by a depthwise-separable convolution:
\begin{equation}
\mathbf D_{2}^{v}=\operatorname{DSConv}_{2}\left(\mathcal U_{2}(\mathbf D_{4}^{v})\right).
\end{equation}

Thus, $\mathbf D_{2}^{v}$ inherits the progressively aggregated semantic information from L23, L15, and L7. Its fusion with fine-scale CNN features is described in the following subsection.

Overall, the DINO hierarchy establishes cross-view correspondences at P16, progressively restores spatial structures at P8 and P4 using L15 and L7, and propagates the aggregated representation to P2. The cross-attended P16 feature is additionally injected into P8 and P4 to maintain coarse-to-fine alignment at these intermediate pyramid levels.
\subsubsection{DINO-guided CNN Detail Enhancement}
Although the multi-level DINO representation maintains stable cross-modal correspondences across scales, its patch-level features struggle to capture fine local structures. CNN features preserve high-frequency cues such as edges, corners, and textures, but are more susceptible to modality gaps and correspondence drift. We therefore use DINO features as semantic anchors and selectively incorporate CNN details to enhance local representations while preserving cross-modal consistency.

Specifically, the CNN branch extracts multi-scale detail features at the P1, P2, P4, and P8 scales, denoted by
$\{\mathbf{E}_s^v\}_{s\in\{1,2,4,8\}}$.
The P8 and P4 features provide local structural information with relatively large receptive fields, while the P2 and P1 features preserve finer edge and texture cues for accurate localization. At the P8 and P4 scales, the DINO features $\mathbf{D}_8^v$ and $\mathbf{D}_4^v$ serve as the primary correspondence representations, while the corresponding CNN features are injected through residual fusion:
\begin{equation}
\widetilde{\mathbf{P}}_s^v
=
T_s(\mathbf{D}_s^v)
+
\alpha_s\Psi_s(\mathbf{E}_s^v),
\qquad s\in\{8,4\},
\end{equation}
where $T_s$ and $\Psi_s$ denote channel projections for the DINO and CNN features, respectively, and $\alpha_s$ is a learnable coefficient controlling the contribution of the CNN details. The fused feature is further combined with the upsampled P16 cross-view feature:
\begin{equation}
\mathbf{P}_s^v
=
M_s\left(
[
\widetilde{\mathbf{P}}_s^v;
\mathcal{U}_{16\rightarrow s}(\mathbf{P}_{16}^v)
]
\right),
\qquad s\in\{8,4\},
\end{equation}
where $\mathcal{U}_{16\rightarrow s}$ denotes upsampling from P16 to scale $s$, and $M_s$ denotes channel fusion. This design supplements the DINO representations with local structures while allowing the coarse-scale cross-modal correspondences to continuously guide subsequent refinement.

At the P2 scale, the higher-resolution CNN features provide more precise structural cues but are also more sensitive to modality-specific appearances. We therefore introduce a conditioned adaptive gate that jointly considers the DINO and CNN features and modulates the CNN details at each spatial location and feature channel before residual injection:
\begin{equation}
\begin{aligned}
\mathbf{A}_2^v
&=
\sigma\left(
\mathcal{G}_2[\mathbf{D}_2^v;\mathbf{E}_2^v]
\right),\\
\mathbf{P}_2^v
&=
\mathbf{D}_2^v
+
\gamma_2\mathbf{A}_2^v\odot\mathbf{E}_2^v,
\end{aligned}
\end{equation}
where $[\cdot;\cdot]$ denotes channel-wise concatenation, $\mathcal{G}_2$ is the gate predictor, $\sigma(\cdot)$ is the sigmoid function, $\mathbf{A}_2^v$ is the adaptive gate, $\odot$ denotes element-wise multiplication, and $\gamma_2$ is a learnable fusion coefficient.

At the P1 scale, a shallow convolutional branch extracts full-resolution structural features, while the upsampled P2 feature provides broader semantic context:
\begin{equation}
\mathbf{P}_1^v
=
S_{\mathrm{full}}(I^v)
+
\beta\,
\mathcal{U}\left(
T_{2\rightarrow1}(\mathbf{P}_2^v)
\right),
\end{equation}
where $S_{\mathrm{full}}$ denotes the shallow convolutional branch, $T_{2\rightarrow1}$ denotes a $1\times1$ projection for channel alignment, and $\beta$ is a learnable fusion coefficient. P1 is retained as a lightweight CNN detail level for limited local refinement.

Although P1 and P2 contain rich high-frequency structures, their modality-sensitive responses are less reliable for directly establishing cross-modal correspondences. In contrast, the DINO-dominated P4 feature maintains stronger cross-modal stability while providing sufficient spatial resolution. Accordingly, P4 serves as the principal representation for fine-scale correspondence estimation, while P1 and P2 mainly provide auxiliary structural cues for limited refinement and bottom-up detail fusion. Specifically, P1 and P2 are first aligned with P4 in spatial resolution and channel dimension, and then concatenated with P4 for detail fusion:
\begin{equation}
\widehat{\mathbf{P}}_4^v
=
\mathbf{P}_4^v
+
\alpha_f
\mathcal{F}_4
\left(
\mathbf{P}_4^v,
\mathbf{P}_2^v,
\mathbf{P}_1^v
\right),
\end{equation}
where $\mathcal{F}_4$ aligns and fuses the detail features, and $\alpha_f$ is a learnable residual weight. The enhanced $\widehat{\mathbf{P}}_4^v$ retains stable DINO correspondences while improving localization with fine structural details.

Through this design, DINO forms the principal correspondence pathway of DCFP, preserving robustness to modality variations, while CNN features provide auxiliary spatial details for fine localization. Rather than treating the two feature types equally, DCFP maintains DINO as the primary correspondence representation and selectively incorporates CNN cues according to the requirements of each scale. Consequently, higher feature resolution no longer sacrifices cross-modal consistency: reliable coarse correspondences are progressively refined into accurate localizations, mitigating refinement drift while balancing correspondence stability and spatial precision.

\subsection{Coarse-to-Fine Matching}
Let $\{\mathbf{F}_s^q,\mathbf{F}_s^r\}$ denote the final DCFP features of the query and reference images at scales $s\in\{16,8,4,2,1\}$, with $\mathbf{F}_4^v=\hat{\mathbf{P}}_4^v$. CDPM first establishes global correspondences at the coarsest P16 level. Using the cross-attended P16 features, we construct the global correlation matrix:
\begin{equation}
\mathbf{C}_{16}
=
\frac{1}{\sqrt{d}}
(\mathbf{F}_{16}^{q})^{\top}
\mathbf{F}_{16}^{r},
\end{equation}
where $d$ denotes the feature dimension. The global correlation matrix is then used to predict the coarse correspondence field $\mathbf{f}_{16}$ and its associated certainty map, which provide the initialization for subsequent refinement.

Starting from $\mathbf{f}_{16}$, the correspondence field is progressively refined at P8, P4, P2, and P1. At each refinement scale, the correspondence field from the preceding pyramid level is first upsampled to the current resolution. The reference features are then sampled according to the current correspondence estimate, and a local correlation volume is constructed around each predicted position. A lightweight refinement module predicts a residual displacement:
\begin{equation}
\begin{aligned}
\Delta\mathbf{f}_{s,t}
&=
\mathcal{R}_{s}
\left(
\mathbf{F}_{s}^{q},
\mathcal{W}
\left(
\mathbf{F}_{s}^{r},
\mathbf{f}_{s,t-1}
\right),
\mathbf{C}_{s,t}
\right),\\
\mathbf{f}_{s,t}
&=
\mathbf{f}_{s,t-1}
+
\Delta\mathbf{f}_{s,t},
\end{aligned}
\end{equation}
where $\mathcal{W}(\cdot)$ denotes reference feature sampling and alignment based on the current correspondence field, $\mathbf{C}_{s,t}$ denotes the local correlation volume, and $\mathcal{R}_{s}$ represents the refinement module at scale $s$. As the feature resolution increases, the local search range is progressively reduced to correct residual localization errors. We perform two refinement iterations at P8 and P4 and one at P2 and P1, concentrating iterative updates on DINO-dominated scales while preserving correspondence stability at fine scales. The final dense flow and certainty map are upsampled to the image resolution, from which reliable correspondences are selected to estimate the homography using RANSAC~\cite{fischler1981random}. At each refinement scale, the network also predicts a certainty map to quantify the reliability of the updated correspondences.

\subsection{Loss Function}\label{sec3.5}
CDPM is optimized in two training phases using separate objectives: a patch-level contrastive loss for DINO adaptation and a multi-scale correspondence loss for matching.

\subsubsection{Patch-level contrastive loss}
Let $\mathbf{z}^{q}{i}$ and $\mathbf{z}^{r}{j}$ denote the normalized tokens of the query and reference images, respectively. For each valid query patch $i$, the augmentation homography $\mathbf{H}_{\mathrm{aug}}$ is used to determine its corresponding reference patch $\pi(i)$. We treat $\mathbf{z}^{r}_{\pi(i)}$ as the positive sample and the remaining reference patches as negative samples. The loss is defined as follows:
\begin{equation}
\mathcal{L}_{\mathrm{ctr}}
=
-\frac{1}{|\mathcal{V}|}
\sum_{i\in\mathcal{V}}
\log
\frac{
\exp\left(
(\mathbf{z}^{q}_{i})^{\top}
\mathbf{z}^{r}_{\pi(i)}
/\tau
\right)
}{
\sum_{j=1}^{N_r}
\exp\left(
(\mathbf{z}^{q}_{i})^{\top}
\mathbf{z}^{r}_{j}
/\tau
\right)
},
\end{equation}
where $\mathcal{V}$ denotes the set of valid query patches, $N_r=784$ is the number of reference patches, and $\tau=0.07$ is the temperature parameter. This loss brings geometrically corresponding cross-modal patches closer in the feature space while separating non-corresponding ones.

\subsubsection{Multi-scale correspondence loss}
For each pyramid scale $s$ and refinement iteration $t$, the network predicts a correspondence field $\hat{\mathbf{p}}^{s,t}$ and a certainty map $\hat{\mathbf{c}}^{s,t}$. The ground-truth correspondence field $\mathbf{p}^{s}$ is obtained by applying the ground-truth homography to the query grid. The correspondence regression loss is defined as
\begin{equation}
\mathcal{L}_{\mathrm{reg}}^{s,t}
=
\frac{1}{|\Omega^{s,t}|}
\sum_{\mathbf{x}\in\Omega^{s,t}}
\rho_s
\left(
\left\|
\hat{\mathbf{p}}^{s,t}(\mathbf{x})
-
\mathbf{p}^{s}(\mathbf{x})
\right\|_2
\right),
\end{equation}
where $\Omega^{s,t}$ is the set of valid locations at scale $s$ and iteration $t$, and $\rho_s(\cdot)$ is a robust regression function defined as:
\begin{equation}
\rho_s(e)
=
(\delta)^{\alpha}
\left[
\left(\frac{e}{\delta}\right)^2+1
\right]^{\alpha/2}.
\end{equation}
The certainty map is supervised using the standard binary cross-entropy loss:
\begin{equation}
\mathcal{L}_{\mathrm{cert}}^{s,t}
=
\operatorname{BCE}
\left(
\hat{\mathbf{c}}^{s,t},
\mathbf{c}^{s}
\right),
\end{equation}
where $\mathbf{c}^{s}$ denotes the ground-truth certainty map. The overall matching objective is
\begin{equation}
\mathcal{L}_{\mathrm{match}}
=
\sum_{s}\sum_{t}
\left(
\mathcal{L}_{\mathrm{reg}}^{s,t}
+
\lambda_{\mathrm{cert}}
\mathcal{L}_{\mathrm{cert}}^{s,t}
\right),
\end{equation}
where $\alpha=0.5$, $\delta=10^{-4}$, and $\lambda_{\mathrm{cert}}=0.01$. At fine scales, supervision is restricted to locations whose ground-truth correspondences fall within the refinement range determined by the preceding coarser prediction.
\begin{figure}[t]
    \centering
    \includegraphics[width=0.49\textwidth]{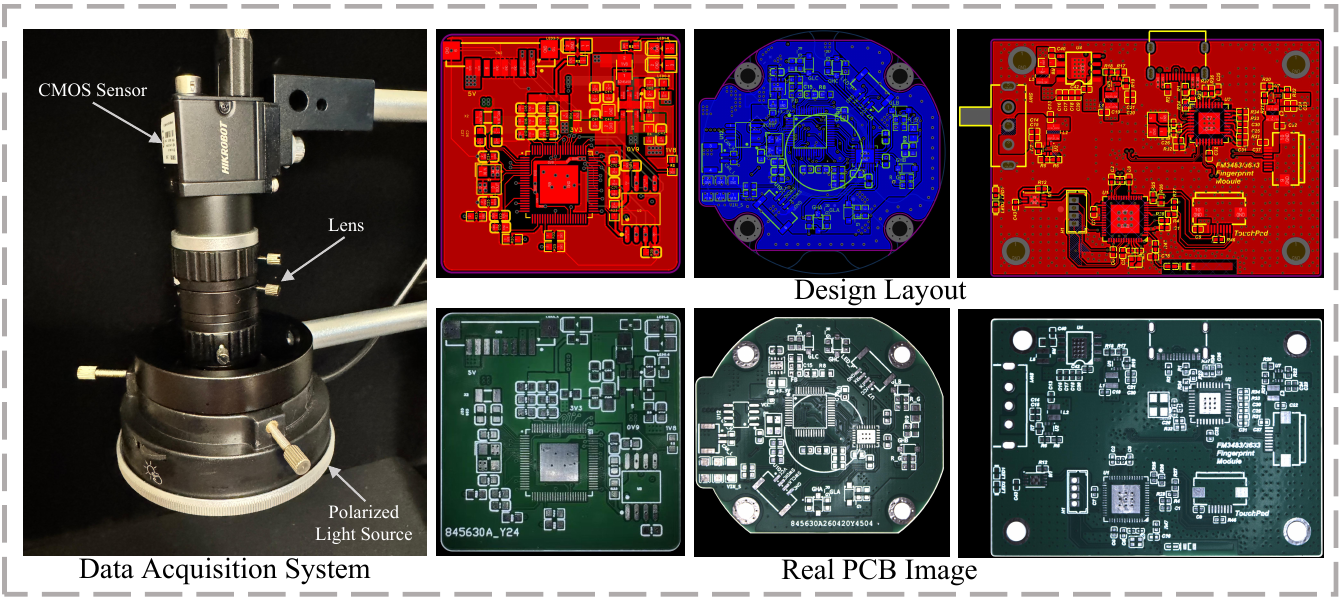}
    \caption{Data acquisition setup and representative PCB-Layout pairs. The imaging system is shown on the left, while representative design layouts and their corresponding real PCB images are shown on the right.}
    \label{PCBLayout}
    \vspace{-0.6cm}
\end{figure}

\begin{table*}[t]
\centering
\caption{Quantitative comparison with state-of-the-art methods on VIS-IR, GoogleMap, and PCB-Layout. AUC@$\{3,5,10,20\}$ ($\uparrow$) uses pixel-error thresholds, and mACE ($\downarrow$) is in pixels. SP+LG denotes SuperPoint with LightGlue, and MatchAny denotes MatchAnything. The best and second-best results are highlighted in \textbf{bold} and \underline{underlined}, respectively.}
\label{QuantitativeComparison}

\footnotesize
\renewcommand{\arraystretch}{1.18}
\setlength{\tabcolsep}{1pt}

\begin{tabular}{@{}
>{\raggedright\arraybackslash}p{2.20cm}
*{15}{C{0.95cm}}
@{}}
\toprule
\multirow[c]{2}{*}{\textbf{Method}}
& \multicolumn{5}{c}{\textbf{VIS-IR}}
& \multicolumn{5}{c}{\textbf{GoogleMap}}
& \multicolumn{5}{c}{\textbf{PCB-Layout}} \\
\cmidrule(lr{3pt}){2-6}
\cmidrule(lr{3pt}){7-11}
\cmidrule(lr{3pt}){12-16}

& @3$px$ & @5$px$  & @10$px$  & @20$px$  & $mACE$
& @3$px$ & @5$px$  & @10$px$  & @20$px$  & $mACE$ 
& @3$px$ & @5$px$  & @10$px$  & @20$px$  & $mACE$ \\
\midrule
SP+LG~\cite{detone2018superpoint,lindenberger2023lightglue}
& 4.48 & 13.53 & 34.16 & 54.91 & 17.22
& 2.71 & 7.98 & 19.72 & 32.27 & 36.39
& 2.86 & 9.31 & 32.59 & 60.63 & 8.11 \\
LoMa~\cite{nordstrom2026loma}
& 2.95  & 12.18 & 32.97 & 54.28 & 16.29
& 0.50  & 2.64  & 10.85 & 25.37 & 32.83
& 5.27  & 12.69 & 34.26 & 61.53 & 8.03 \\
\addlinespace[3pt]
RHWF~\cite{cao2023recurrent}
& 19.15 & 33.79 & 56.86 & 74.95 & 6.23
& 5.46  & 17.73 & 42.42 & 63.35 & 10.72
& 0.00  & 0.00  & 0.81  & 13.56 & 20.47 \\

MCNet~\cite{zhu2024mcnet}
& 20.90 & 34.99 & 56.36 & 73.97 & 6.71
& 18.77 & 34.59 & 54.12 & 68.35 & 10.27
& 0.11  & 0.77  & 14.64 & 41.41 & 14.80 \\
\addlinespace[3pt]
LoFTR~\cite{sun2021loftr}
& 7.92  & 19.63 & 40.73 & 59.62 & 13.99
& 6.54  & 18.28 & 41.43 & 61.27 & 11.59
& 4.23  & 16.28 & 43.56 & 67.19 & 8.26 \\

XoFTR~\cite{tuzcuouglu2024xoftr}
& 12.34 & 23.90 & 43.27 & 61.12 & 14.15
& 10.21 & 22.77 & 44.01 & 62.68 & 12.24
& 3.36  & 14.71 & 39.68 & 64.78 & 7.49 \\
ELoFTR~\cite{wang2024eloftr}
& 17.64 & 31.26 & 52.72 & 70.16 & 9.20
& 5.42  & 14.59 & 34.03 & 54.27 & 13.57
& 0.11  & 1.02  & 14.61 & 40.42 & 17.17 \\

ELoFTR$_{\scalebox{0.8}{$\scriptscriptstyle\mathrm{MINIMA}$}}$~\cite{ren2025minima}
& 2.57  & 9.81  & 28.96 & 50.72 & 17.45
& 1.27  & 5.17  & 15.23 & 29.02 & 34.59
& 0.00  & 0.38  & 7.36  & 37.17 & 13.62 \\

ELoFTR$_{\scalebox{0.8}{$\scriptscriptstyle\mathrm{MatchAny}$}}$~\cite{he2025matchanything}
& 3.91  & 12.52 & 31.87 & 51.54 & 18.80
& 0.25  & 0.86  & 3.00  & 6.43  & 60.55
& 2.75  & 13.61 & 42.28 & 67.70 & 7.04 \\
\addlinespace[3pt]
RoMa~\cite{edstedt2024roma}
& \underline{34.07} & \underline{45.42} & 62.74 & 76.75 & 5.83
& 33.57 & 49.94 & 68.53 & 81.60 & 4.48
& 5.25 & 24.82 & \underline{55.86} & \underline{76.44} & \underline{4.83} \\

RoMa$_{\scriptscriptstyle\mathrm{MINIMA}}$~\cite{ren2025minima}
& 5.52  & 17.08 & 39.32 & 60.61 & 12.81
& 5.02  & 11.47 & 23.83 & 36.79 & 30.64
& 9.92  & \underline{28.39} & 54.63 & 75.34 & 4.98 \\

RoMa$_{\scriptscriptstyle\mathrm{MatchAny}}$~\cite{he2025matchanything}
& 5.65  & 17.55 & 39.40 & 60.29 & 12.27
& 3.57  & 10.50 & 25.02 & 41.58 & 23.82
& 6.26  & 24.38 & 53.99 & 74.16 & 5.53 \\

RoMa v2~\cite{edstedt2025roma}
& 20.82 & 39.92 & \underline{63.99} & \underline{80.34} & \underline{3.62}
& 21.44 & 40.21 & 63.14 & 79.51 & 4.65
& 1.59 & 7.65 & 33.16 & 62.54 & 7.58 \\
GFNet~\cite{zhang2025adapting}
& 21.05 & 35.56 & 58.29 & 76.15 & 5.62
& \underline{51.28} & \underline{66.13} & \underline{80.50} & \underline{89.16} & \underline{2.65}
& \underline{10.44} & 25.22 & 52.67 & 74.50 & 5.27 \\
\midrule
\textbf{Ours}
& \textbf{41.43} & \textbf{58.82} & \textbf{76.49} & \textbf{87.17} & \textbf{2.78}
& \textbf{52.37} & \textbf{67.71} & \textbf{81.85} & \textbf{89.91} & \textbf{2.50}
& \textbf{15.61} & \textbf{36.63} & \textbf{63.33} & \textbf{80.38} & \textbf{3.95} \\
\bottomrule
\end{tabular}

\vspace{1.5mm}
\begin{minipage}{\textwidth}
\end{minipage}
\vspace{-0.65cm}
\end{table*}
\section{Experiments}\label{sec5}
\subsection{Datasets and Evaluation Protocol}
We evaluate our method on three multimodal datasets: VIS-IR, GoogleMap, and our PCB-Layout dataset. The VIS-IR and GoogleMap datasets, provided by~\cite{zhang2025adapting}, are used for visible–infrared and satellite–map matching, respectively. VIS-IR contains 17,119 training image pairs at $640 \times 512$, whereas GoogleMap contains 6,048 training image pairs at $1280 \times 1280$; both additionally include 1,000 test image pairs.
We constructed PCB-Layout using real PCB images and layout diagrams exported from EDA software. As illustrated in~\autoref{PCBLayout}, it includes diverse PCB-Layout pairs and the corresponding acquisition system. Only task-relevant layers were retained, and back-side layouts were flipped to ensure geometric consistency. PCB images were captured at $3072 \times 2048$ using a HIKROBOT MV-CU060-10UC industrial camera with a 10--50\,mm zoom lens and polarized illumination. Each pair was annotated with a homography, whose accuracy was verified by manual inspection and reprojection error. The dataset comprises 100 PCB surfaces, with seven images per surface captured under dark, normal, and bright illumination, yielding 700 images. The surfaces were split 8:2 into training and test sets. After cropping and augmentation, 3,000 training pairs and 500 test pairs were obtained.

We evaluate registration performance using mACE and AUC@${3,5,10,20}$, where higher AUC and lower mACE indicate better accuracy. For correspondence-based methods, we retain the top 5,000 matches by confidence and estimate the homography using RANSAC~\cite{fischler1981random}, while direct methods use their predicted homographies.

\subsection{Implementation Details}
All images are resized to $448\times448$. For a fair comparison, all baselines are retrained on the training sets using official implementations, except variants using MINIMA~\cite{ren2025minima} or MatchAnything~\cite{he2025matchanything} weights, which directly adopt the official weights. During patch-level DINO adaptation, we optimize the Q/K/V projections in Layers 8--15 and 16--23 sequentially, with 4,000 iterations for each stage. The adaptation is performed separately for each dataset using its own training set.  AdamW~\cite{Loshchilov2017DecoupledWD} is used with a batch size of 6, a fixed learning rate of $5\times10^{-5}$, and a contrastive temperature of 0.07. The downstream matching network is trained for up to 80 epochs with a batch size of 12 using AdamW. The initial learning rate is $1.5\times10^{-4}$ and is decayed using a cosine schedule, with a weight decay of 0.01. The gradient norm is clipped to 1.0. After the two-stage adaptation, the DINOv3 backbone is kept frozen during downstream matching training, while only the feature pyramid and refinement modules are optimized. All experiments are conducted using PyTorch 2.2.0 on a single NVIDIA GeForce RTX 4090 GPU with 24\,GB of memory, hosted by an Intel Core i7-13700KF CPU with 64\,GB of RAM under Ubuntu 20.04.
\begin{figure*}[t]
    \centering

    \subfloat[Registration comparison.\label{Transform}]{
        \includegraphics[width=\textwidth]{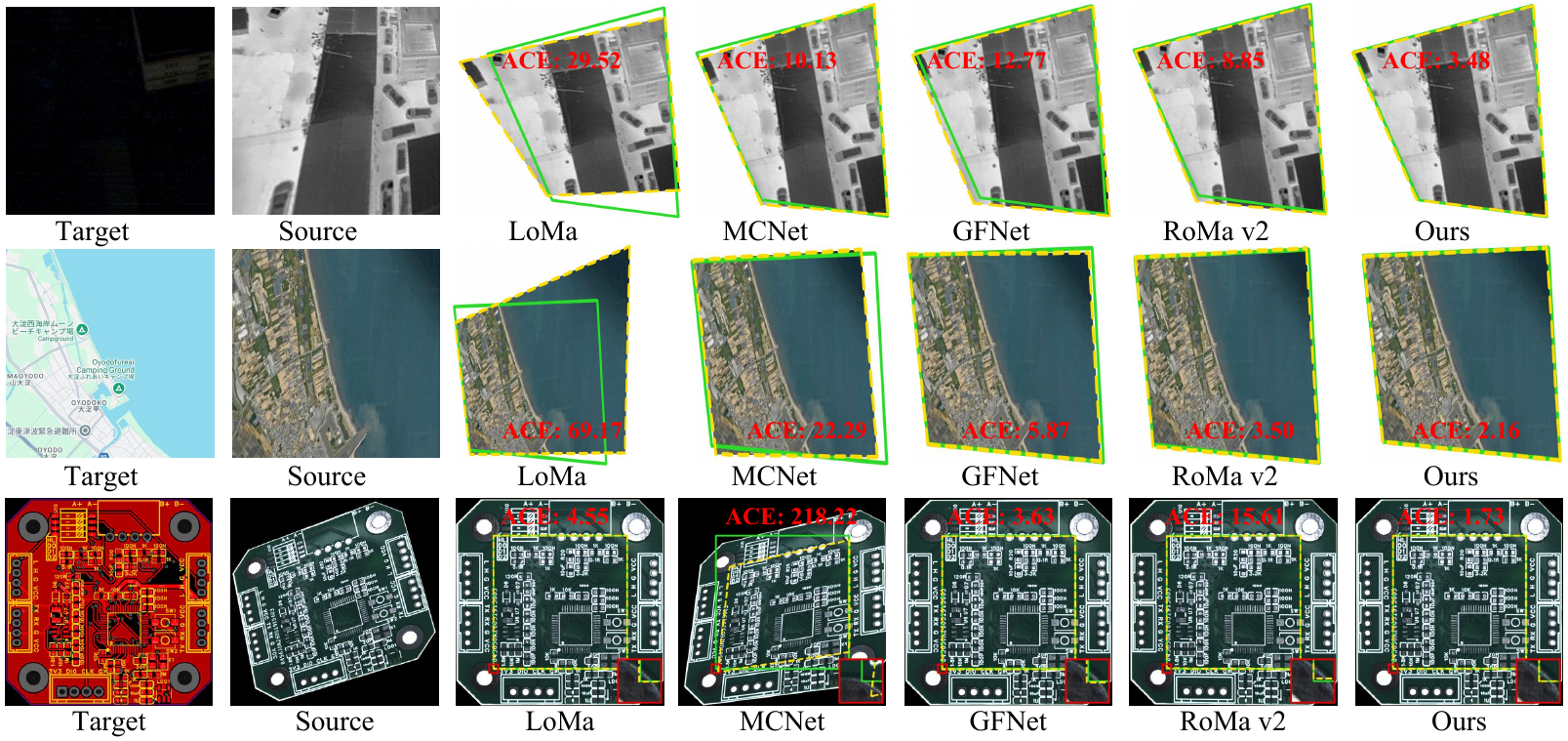}
    }

    \vspace{-0.05cm}

    \subfloat[Correspondence comparison.\label{MatchLine}]{
        \includegraphics[width=\textwidth]{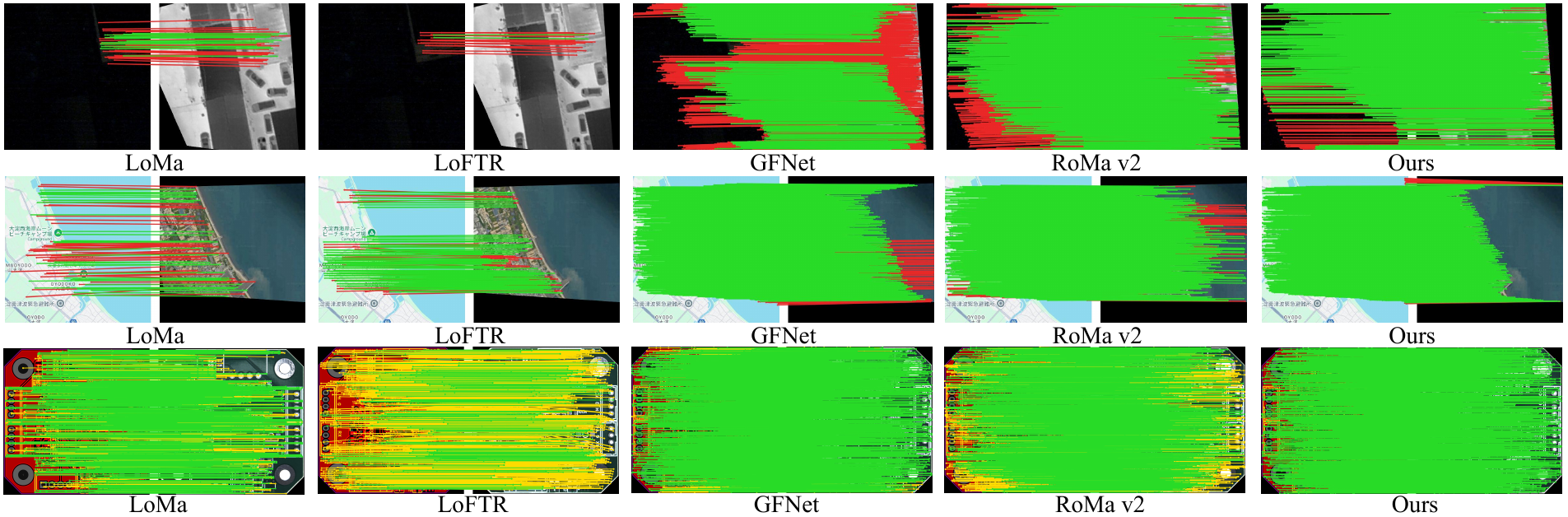}
    }

    \vspace{-0.1cm}
    \caption{Qualitative comparison on VIS-IR, GoogleMap, and PCB-Layout. 
    (a) Alignment results. Green solid and yellow dashed boundaries 
    denote the ground truth and prediction, respectively. 
    (b) Correspondence results. Green and red/yellow lines indicate reprojection 
    errors of $\leq 3$ and $>3$ pixels.}
    \label{fig:qualitative}
    \vspace{-0.65cm}
\end{figure*}
\subsection{Comparison with State-of-the-Art Methods}
We compare CDPM with  three paradigms of methods. The first category comprises direct homography estimation methods, including RHWF~\cite{cao2023recurrent} and MCNet~\cite{zhu2024mcnet}. The second category includes sparse and semi-dense matching methods, including LoMa~\cite{nordstrom2026loma}, SuperPoint~\cite{detone2018superpoint} with LightGlue~\cite{lindenberger2023lightglue}, LoFTR~\cite{sun2021loftr}, XoFTR~\cite{tuzcuouglu2024xoftr}, and ELoFTR~\cite{wang2024eloftr}. The third category consists of dense matching methods, including GFNet~\cite{zhang2025adapting}, RoMa~\cite{edstedt2024roma}, and RoMa v2~\cite{edstedt2025roma}. Notably, MINIMA~\cite{ren2025minima} and MatchAnything~\cite{he2025matchanything} are general cross-modal training frameworks. The quantitative results are reported in~\autoref{QuantitativeComparison}.

On the VIS-IR dataset, CDPM achieves the best performance across all evaluation metrics. Among existing approaches, dense matching methods are generally more competitive: RoMa~\cite{edstedt2024roma} ranks second on AUC@3 and AUC@5, while RoMa v2~\cite{edstedt2025roma} achieves the second-best results on AUC@10, AUC@20, and mACE. RoMa v2~\cite{edstedt2025roma} improves overall robustness but exhibits a notable drop in AUC@3, suggesting that fine-scale localization remains more sensitive to cross-modal appearance variations under strict thresholds. Compared with the corresponding second-best methods, CDPM improves the four AUC scores by 7.36, 13.40, 12.50, and 6.83 percentage points, corresponding to relative gains of 21.6\%, 29.5\%, 19.5\%, and 8.5\%, respectively. The substantial improvements under the strict 3- and 5-pixel thresholds indicate that CDPM can constrain the registration errors of more test samples within a small range. Meanwhile, CDPM reduces mACE from 3.62 to 2.78 pixels, representing a 23.2\% reduction and demonstrating higher registration accuracy over the overall error distribution. Compared with direct homography estimation and sparse or semi-dense matching methods, CDPM exhibits more pronounced advantages, validating the importance of preserving high-level semantic consistency and progressively recovering fine-grained geometric structures when handling the substantial modality gap between visible and infrared images. In addition, although the variants trained with MINIMA~\cite{ren2025minima} and MatchAnything~\cite{he2025matchanything} exhibit strong cross-modal generalization, they are less effective at preserving precise geometric correspondences.
\begin{figure*}[t]
    \centering
    \includegraphics[width=1.0\textwidth]{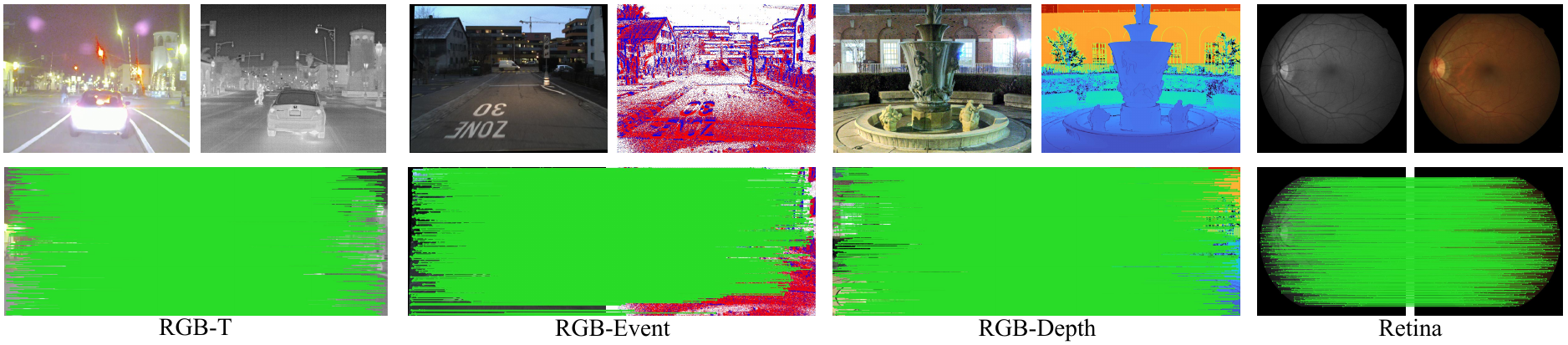}
    \vspace{-0.6cm}
    \caption{Qualitative results of the transferability experiment on RGB-T, RGB-Event, RGB-Depth, and Retina datasets.}
    \label{Generalize}
    \vspace{-0.6cm}
\end{figure*}

GoogleMap and PCB-Layout both involve cross-representation matching between real images and digital planar diagrams, where the former matches satellite imagery with digital maps, while the latter matches real PCB images with EDA layout diagrams. Both datasets preserve relatively stable planar geometry but exhibit substantial differences in texture, color, and visual representation, thereby providing a suitable test of whether a model can maintain geometric consistency across large appearance variations. On GoogleMap, CDPM achieves AUC@${3,5,10,20}$ scores of 52.37, 67.71, 81.85, and 89.91, respectively, outperforming the second-best method, GFNet~\cite{zhang2025adapting}, by 1.09, 1.58, 1.35, and 0.75, respectively, while reducing mACE from 2.65 to 2.50 pixels. These results demonstrate that CDPM can preserve stable and accurate geometric correspondences despite the substantial representational gap between satellite imagery and digital maps. On the more challenging PCB-Layout dataset, CDPM achieves AUC scores of 15.61, 36.63, 63.33, and 80.38, surpassing the corresponding second-best results by 5.17, 8.24, 7.47, and 3.94 percentage points, respectively, while reducing mACE from 4.83 to 3.95 pixels. Overall, CDPM establishes reliable and precise planar geometric correspondences in scenarios involving substantial modality differences between real images and digital planar diagrams.

To provide a more intuitive comparison of different matching and registration methods,~\autoref{Transform} and~\autoref{MatchLine} present the homography registration results and correspondence visualizations, respectively. LoMa achieves reasonable registration on the PCB-Layout dataset, but exhibits noticeable boundary deviations on VIS-IR and GoogleMap, indicating that sparse correspondences are easily limited by matching quality under substantial modality gaps. MCNet~\cite{zhu2024mcnet} shows large geometric deviations on GoogleMap and PCB-Layout, suggesting that directly regressing homographies is less robust to complex cross-modal appearance and structural variations. GFNet~\cite{zhang2025adapting} performs well overall, but still suffers from evident errors in the low-light VIS-IR scene. RoMa v2~\cite{edstedt2025roma} remains relatively stable across all three datasets, yet its overall accuracy is still lower than that of CDPM. In contrast, CDPM consistently achieves more accurate geometric alignment across different tasks and scenarios, while maintaining strong geometric consistency throughout, with the lowest ACE.

The correspondence visualizations further show that LoMa~\cite{nordstrom2026loma} and LoFTR~\cite{sun2021loftr} produce more erroneous matches under substantial modality differences. Although GFNet~\cite{zhang2025adapting} and RoMa v2~\cite{edstedt2025roma} establish denser correspondences, anomalous matches still occur in low-light and weak-texture regions, areas with large cross-modal appearance discrepancies, and repetitive structures. CDPM maintains more continuous, dense, and geometrically consistent correspondences in these challenging regions while effectively suppressing mismatches.
\begin{table}[t]
    \centering
    \caption{Comparison of model complexity and inference efficiency at an input resolution of $448\times448$.}
    \label{tab:complexity_comparison}
    
    \footnotesize
    \renewcommand{\arraystretch}{1.12}
    \setlength{\tabcolsep}{3pt}
    
    \begin{tabular*}{\columnwidth}{
        @{\extracolsep{\fill}}
        lcccc
        @{}
    }
        \toprule
        Method & Input Size & Params (M) $\downarrow$ & FLOPs (G) $\downarrow$ & Time (ms) $\downarrow$ \\
        \midrule
        SP+LG~\cite{detone2018superpoint,lindenberger2023lightglue}         & $448\times448$ & 13.15 & 317.50 & $23.34\pm0.20$ \\
        LoFTR~\cite{sun2021loftr}         & $448\times448$ & 11.56 & 467.97 & $19.36\pm0.11$ \\
        XoFTR~\cite{tuzcuouglu2024xoftr}         & $448\times448$ & 11.09 & 281.15 & $19.17\pm0.11$ \\
        MCNet~\cite{zhu2024mcnet}         & $448\times448$ & 0.85 & 112.68 & $25.52\pm0.94$ \\
        RoMa~\cite{edstedt2024roma}         & $448\times448$ & 415.67 & 2100.00 & $245.00\pm0.28$ \\
        GFNet~\cite{zhang2025adapting}         & $448\times448$ & 308.24 & 1514.30 & $81.81\pm2.33$ \\
        RoMa v2~\cite{edstedt2025roma}        & $448\times448$ & 425.42 & 1847.80 & $206.59\pm0.26$ \\
        \textbf{Ours} & $448\times448$ & 307.27 & 1004.55 & $74.25\pm1.70$ \\
        \bottomrule
        \vspace{-0.6cm}
    \end{tabular*}
\end{table}
\subsection{Cross-modal Transferability}
To evaluate the cross-modal transferability of CDPM across diverse unseen settings, we directly apply the same model trained on GoogleMap to unseen scenarios and modalities, including RGB-T street scenes~\cite{Xu2020FusionDNAU, tang2022piafusion}, RGB-Event~\cite{Gehrig21ral}, RGB-Depth~\cite{li2018megadepth}, and Retina~\cite{ma2019locality}, without fine-tuning. As shown in~\autoref{Generalize}, CDPM establishes geometrically consistent correspondences under challenging illumination in RGB-T scenes and across different modalities with substantially different imaging characteristics. These qualitative examples show that CDPM can establish geometrically coherent correspondences across multiple previously unseen modality combinations. This transferability benefits from preserving high-level semantic cues throughout the feature pyramid.

\subsection{Complexity Discussion}
Table~\ref{tab:complexity_comparison} compares the model complexity and efficiency of different methods under identical settings and the same input resolution. CDPM contains 307.27M parameters, requires 1004.55G FLOPs, and achieves an average inference time of 74.25 ms. CDPM employs DINO features as the primary representation for cross-modal matching and incorporates a lightweight convolutional network to complement fine-grained structural information, thereby improving localization accuracy while limiting additional computational overhead. Compared with GFNet~\cite{zhang2025adapting}, CDPM reduces FLOPs by 33.7\% and inference time by 9.2\%. Compared with the dense matching methods RoMa~\cite{edstedt2024roma} and RoMa v2~\cite{edstedt2025roma}, CDPM reduces the parameter count by 26.1\% and 27.8\%, FLOPs by 52.2\% and 45.6\%, and inference time by 69.7\% and 64.1\%, respectively. Together, these results show that CDPM achieves superior registration accuracy while requiring lower FLOPs and inference time than all compared dense matching methods.
\begin{figure*}[t]
    \centering
    \includegraphics[width=0.98\textwidth]{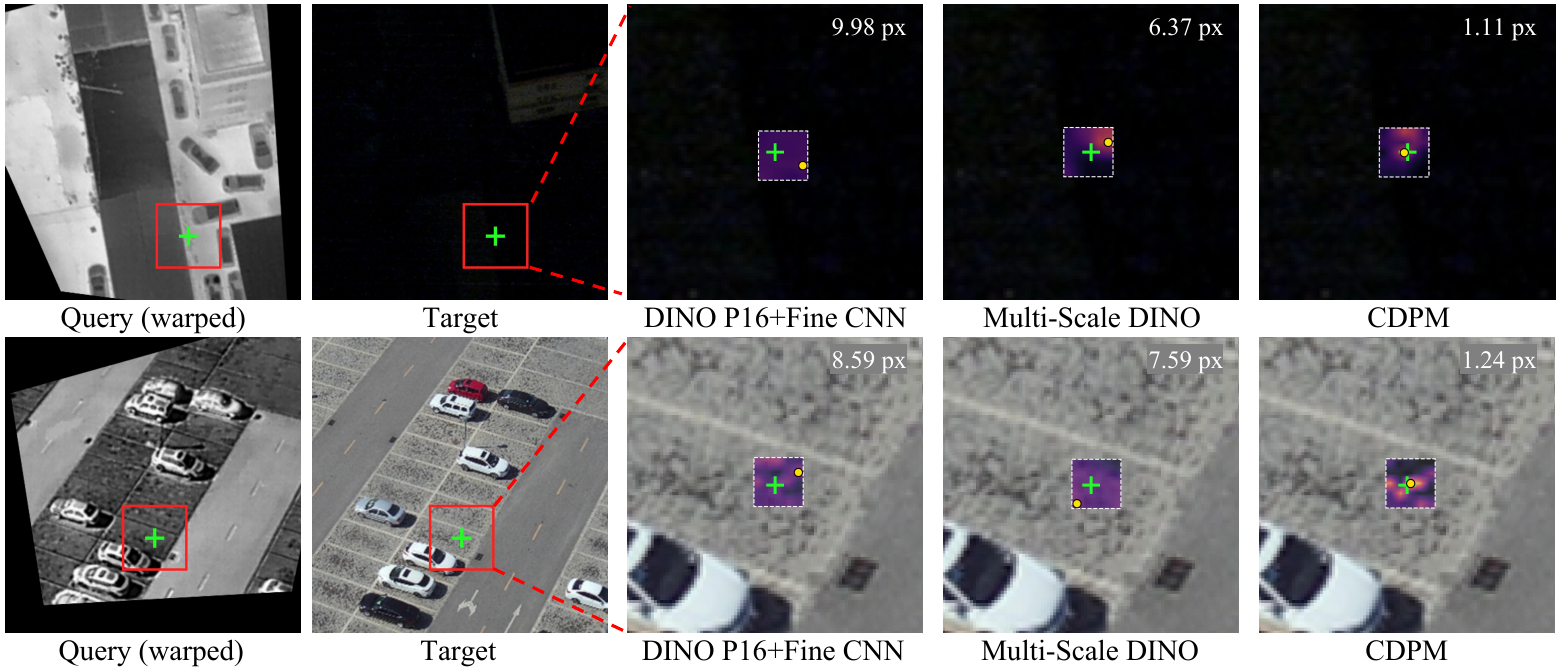}
    \vspace{-0.3cm}
    \caption{Qualitative ablation of DINO-P16 with CNN features at P8--P1, multi-scale DINO without CNN features, and the full CDPM. Green crosses and colored dots indicate the ground-truth and predicted correspondences, respectively, with localization errors reported in pixels.}
    \label{Ablation}
    \vspace{-0.6cm}
\end{figure*}
\begin{table}[t]
    \centering
    \caption{We evaluate cross-modal patch retrieval using Recall@K (R@K) and median rank (MedRank), where R@K measures top-$K$ retrieval accuracy, while MedRank denotes the median rank of the ground-truth correspondence.}
    \label{Dino_Abalation}
    
    \footnotesize
    \renewcommand{\arraystretch}{1.12}
    \setlength{\tabcolsep}{3pt}
    
    \begin{tabular*}{\columnwidth}{
        @{\extracolsep{\fill}}
        lcccc
        @{}
        }
        \toprule
        Configuration 
        & R@1 $\uparrow$ & R@5 $\uparrow$ & R@20 $\uparrow$ & MedRank $\downarrow$ \\
        \midrule
        Raw DINOv3
        & 0.109 & 0.338 & 0.656 & \underline{11} \\
        
        Joint Adaptation
        & \underline{0.581} & \underline{0.941} & \underline{0.989} & \textbf{1} \\
        
        Staged Adaptation
        & \textbf{0.608} & \textbf{0.951} & \textbf{0.991} & \textbf{1} \\
        \bottomrule
        \vspace{-0.7cm}
    \end{tabular*}
\end{table}

\begin{table}[t]
    \centering
    \caption{Ablation study of DINO adaptation, feature construction, and refinement strategies on the VIS-IR dataset.}
    \label{Architecture_Ablation}

    \footnotesize
    \renewcommand{\arraystretch}{1.02}
    \setlength{\tabcolsep}{3pt}

    \begin{tabularx}{\columnwidth}{@{}
        >{\raggedright\arraybackslash}X
        ccccc
    @{}}
    \toprule
    Configuration
    & AUC@3 $\uparrow$
    & @5 $\uparrow$
    & @10 $\uparrow$
    & @20 $\uparrow$
    & mACE $\downarrow$ \\
    \midrule
    \rowcolor{gray!20}
    \multicolumn{6}{c}{\textit{DINO Adaptation}} \\
    \cmidrule(lr){1-6}
    Raw DINOv3
    & 35.03 & 52.33 & 71.56 & 84.19 & 3.62 \\
    E2E Adaptation
    & 36.43 & 53.85 & 73.22 & 85.35 & 3.19 \\
    Joint Adaptation
    & \underline{37.19} & \underline{54.55} & \underline{73.65}
    & \underline{85.55} & \underline{3.15} \\
    Staged Adaptation
    & \textbf{41.43} & \textbf{58.82} & \textbf{76.49}
    & \textbf{87.17} & \textbf{2.78} \\

    \addlinespace[4pt]
    \midrule
    \rowcolor{gray!20}
    \multicolumn{6}{c}{\textit{Feature Construction}} \\
    \cmidrule(lr){1-6}
    DINO-P16+FineCNN
    & 33.51 & 51.39 & 71.49 & 84.33 & 3.40 \\
    Multi-scale DINO
    & \underline{35.12} & \underline{54.13} & \underline{73.69}
    & \underline{85.65} & \underline{3.12} \\
    DCFP
    & \textbf{41.43} & \textbf{58.82} & \textbf{76.49}
    & \textbf{87.17} & \textbf{2.78} \\

    \addlinespace[4pt]
    \rowcolor{gray!20}
    \multicolumn{6}{c}{\textit{Refinement Strategy}} \\
    \cmidrule(lr){1-6}
    $1\times$ Refinement
    & \underline{39.92} & 57.22 & 75.27 & 86.51 & 2.98 \\
    $2\times$ Refinement
    & 39.67 & \underline{57.33} & \underline{75.55}
    & \underline{86.54} & \underline{2.92} \\
    P8--4 ($2\times$), P2--1 ($1\times$)
    & \textbf{41.43} & \textbf{58.82} & \textbf{76.49}
    & \textbf{87.17} & \textbf{2.78} \\

    \bottomrule
    \end{tabularx}
    \vspace{-0.6cm}
\end{table}

\subsection{Ablation Study}
To evaluate the effectiveness of each component, we conduct ablation studies on the VIS-IR dataset covering DINO adaptation, feature construction, and refinement strategy. Cross-modal patch retrieval is additionally used to assess the representation-level correspondence ability under different DINO adaptation strategies. Within each ablation group, all other components are kept unchanged to isolate the effect of the evaluated design.

\subsubsection{Effect of DINO Adaptation}
As shown in~\autoref{Dino_Abalation}, DINO adaptation substantially improves cross-modal feature correspondence. Raw DINOv3 achieves only an R@1 of 0.109 and a MedRank of 11, consistent with the ambiguous cross-modal responses observed in~\autoref{Motivation}. Joint adaptation increases R@1 to 0.581 and reduces MedRank to 1, while staged adaptation further improves R@1, R@5, and R@20 to 0.608, 0.951, and 0.991, respectively. These results indicate that progressive optimization more effectively reduces the cross-modal representation gap and promotes ground-truth correspondences to higher ranks. The representation-level improvements also translate into better end-to-end registration performance. As shown in~\autoref{Architecture_Ablation}, we compare frozen DINO, end-to-end (E2E) adaptation, joint adaptation, and staged adaptation under the same registration pipeline. E2E adaptation jointly optimizes DINO and the matcher in an end-to-end manner, reducing mACE from 3.62 to 3.19 pixels. Joint adaptation further reduces mACE to 3.15 pixels, while staged adaptation achieves the best performance with an mACE of 2.78 pixels and AUC scores of 41.43, 58.82, 76.49, and 87.17. These results show that downstream supervision can improve DINO representations, but patch-level adaptation provides more direct geometric supervision, allowing DINO features to better reflect true cross-modal correspondences.

\subsubsection{Effect of DCFP and Refinement}
As shown in~\autoref{Architecture_Ablation}, the architectural ablations reveal clear differences among feature construction and refinement strategies. Using DINO-P16 with CNN features at P8--P1 achieves AUC@${3,5,10,20}$ of 33.51, 51.39, 71.49, and 84.33, respectively, with an mACE of 3.40 pixels. Replacing these fine-scale CNN features with multi-scale DINO improves the corresponding scores to 35.12, 54.13, 73.69, and 85.65, while reducing mACE to 3.12 pixels. The complete DCFP further reduces mACE to 2.78 pixels by selectively incorporating auxiliary CNN details. This consistent improvement indicates that preserving DINO representations across scales is important for maintaining reliable cross-modal correspondence. On this basis, refinement improves the overall alignment accuracy, but the gains are not consistent across scales. Using one refinement iteration at each scale achieves AUC scores of 39.92, 57.22, 75.27, and 86.51, while reducing mACE to 2.98 pixels. Increasing to two iterations slightly reduces mACE to 2.92 pixels but decreases AUC@3 to 39.67, suggesting that excessive fine-scale refinement may disturb reliable correspondences. Using two refinement iterations at P8--P4 and only one at P2/P1 achieves the best overall performance, with an AUC@3 of 41.43 and an mACE of 2.78 pixels.

The qualitative results in~\autoref{Ablation} illustrate the behavior of different feature constructions under challenging conditions. 
In the first low-light scene, DINO-P16 with CNN features at P8--P1 produces no clear response around the ground-truth correspondence, suggesting that the semantic correspondence encoded by DINO is weakened during fine-scale matching. Multi-scale DINO preserves the correct response, but its activation remains relatively diffuse. In contrast, CDPM retains the semantic correspondence while producing a more concentrated local response, reducing the localization error from 9.98 to 1.11 pixels. 
In the second scene with repetitive textures, both DINO-P16 with CNN features and multi-scale DINO exhibit ambiguous responses among similar structures, whereas CDPM successfully identifies the correct correspondence and reduces the error from 8.59 to 1.24 pixels. These results show that CDPM better preserves semantic consistency while enhancing local discriminability, leading to more reliable fine-scale localization.

Overall, these results validate the DINO-centric design of DCFP. Maintaining DINO representations across scales preserves reliable cross-modal correspondence, while selectively incorporating CNN details improves local discriminability without overwhelming the semantic cues. This balance enables accurate and robust fine-scale localization.

\section{Conclusion}
In this paper, we propose CDPM for cross-modal image matching and planar registration to jointly achieve stable correspondences and accurate localization under substantial modality differences. First, to bridge the gap between semantic similarity and geometric correspondence, we specifically adapt DINOv3 using geometrically consistent cross-modal patch pairs, enabling its semantic representations to better reflect true spatial correspondences. Second, to preserve correspondence stability during iterative fine-scale refinement, we further develop a DINO-Centric Feature Pyramid that preserves stable correspondences through multi-scale DINO representations while introducing a lightweight CNN branch for complementary structural details. Through this framework, CDPM achieves accurate fine-grained localization while maintaining stable cross-modal semantic correspondences, delivering superior registration performance across multiple cross-modal datasets while requiring fewer FLOPs and lower inference time than representative dense matchers.

\section*{Acknowledgments}
This work was supported in part by the National Natural Science Foundation of China under Grant U24A20270, by the Zhejiang Province Leading Geese Plan under Grant 2025C02013, and by the Innovation and Technology Fund of Hong Kong, China, under Grant ITS/341/23.

\bibliographystyle{IEEEtran}
\bibliography{refs}

\end{document}